\documentclass[11pt]{article}

\usepackage[a4paper,margin=1in]{geometry}
\usepackage{setspace}
\usepackage{graphicx}
\usepackage{amsmath, amssymb}
\usepackage{booktabs}
\usepackage{array}
\usepackage{hyperref}
\usepackage{caption}
\usepackage{subcaption}
\usepackage{enumitem}
\usepackage{float}
\usepackage{cleveref}
\usepackage{microtype}
\usepackage{tabularx}
\usepackage{ragged2e}
\usepackage{placeins}
\usepackage{needspace}
\usepackage{caption}   

\usepackage[T1]{fontenc}
\usepackage{lmodern}

\newcolumntype{Y}{>{\RaggedRight\arraybackslash}X}

\title{\textbf{An Architecture-Led Hybrid Report on Vision-Language Models for Frame-Level Person Detection and Emotion-Oriented Attribute Extraction}}
\author{Thomson Tong \and Diba Darooneh}
\date{December 25, 2025}

\begin{document}
\maketitle

\begin{abstract}
This report provides an architecture-led analysis of two modern vision-language models (VLMs)-\textit{Qwen2.5-VL-7B-Instruct} and \textit{Llama-4-Scout-17B-16E-Instruct}-and explains how their architectural properties map to a practical video-to-artifact pipeline implemented in the BodyLanguageDetection repository \cite{bodylanguagedetection_repo}. The system samples video frames, prompts a VLM to detect visible people and generate pixel-space bounding boxes with prompt-conditioned attributes (emotion by default), validates output structure using a predefined schema, and optionally renders an annotated video. We first summarize the shared multimodal foundation (visual tokenization, Transformer attention, and instruction following), then describe each model’s architecture at a level sufficient to justify engineering choices without speculative internals. Finally, we connect model behavior to system constraints: structured outputs can be syntactically valid while semantically incorrect, schema validation is structural (not geometric correctness), person identifiers are frame-local in the current prompting contract, and interactive single-frame analysis returns free-form text rather than schema-enforced JSON. These distinctions are critical for writing defensible claims, designing robust interfaces, and planning evaluation.
\end{abstract}


\section{Introduction}

A common failure mode in applied machine learning writeups is to describe a system pipeline without clearly explaining what the model can and cannot guarantee. This is especially risky for vision-language models (VLMs): they can produce convincing structured outputs while still being wrong about what is in an image, where it is located, or which attribute should be assigned. A defensible report must therefore be architecture-led: it should explain what the model is doing ``under the hood'' and then translate that into system-level interface decisions and limitations.

The BodyLanguageDetection project \cite{bodylanguagedetection_repo} uses VLMs to convert sampled video frames into machine-consumable artifacts: per-frame person detections, pixel-space bounding boxes, and prompt-conditioned per-person attributes (emotion by default), plus an optional annotated video. Two models play different roles: Qwen2.5-VL-7B-Instruct is used for batch, schema-oriented extraction; Llama-4-Scout-17B-16E-Instruct is used for interactive, single-frame inspection and prompt iteration.

The contributions of this report are:
\begin{itemize}
    \item An architecture-led explanation of how multimodal Transformers represent images as tokens and fuse them with text instructions using attention \cite{vaswani2017attention,dosovitskiy2021vit};
    \item A model-specific discussion of Qwen2.5-VL-7B-Instruct and Llama-4-Scout-17B-16E-Instruct grounded in public documentation \cite{qwen2_5_vl_model_card,nvidia_llama4_scout_nim,google_vertex_llama4_scout};
    \item A system mapping that makes interface claims precise: what is validated, what is only prompted, what can fail silently, and what is not implemented (e.g., cross-frame identity tracking).
\end{itemize}

\section{System Overview}

This section summarizes only the system context needed to motivate architectural choices. The BodyLanguageDetection pipeline \cite{bodylanguagedetection_repo} transforms an uploaded video into structured artifacts through four stages:

\begin{enumerate}[leftmargin=*]
    \item \textbf{Frame sampling:} Videos are not processed at native frame rate. Instead, frames are sampled at a configurable interval using OpenCV \cite{opencv_videocapture_tutorial}. Sampling reduces cost and payload size while preserving semantic coverage for cues that change more slowly than the native frame rate.
    \item \textbf{Batch VLM inference (primary artifacts):} Sampled frames are sent to \textit{Qwen2.5-VL-7B-Instruct} using a chat-completions style endpoint \cite{hf_chat_completion_docs}. The model is instructed to detect all visible people in each frame and produce one JSON object per batch containing per-frame detections, pixel-space bounding boxes, confidence, and prompt-conditioned attributes.
    \item \textbf{Structural validation and artifact writing:} The system parses the model output as JSON and validates structure using a predefined schema (Pydantic) \cite{pydantic_docs_overview}. This step enforces field presence/types/ranges but does \textit{not} guarantee semantic correctness (e.g., it cannot prove that a box tightly encloses a person).
    \item \textbf{Video annotation:} A downstream annotator consumes the detections artifact and draws bounding boxes/labels onto the video to produce an annotated MP4. This is a visualization step; it should be treated as rendering model hypotheses, not ground truth.
\end{enumerate}

\needspace{7\baselineskip}
\textbf{Implementation assumptions and guarantees:} The current prompting contract assigns \texttt{person\_id} values uniquely \emph{within a frame} (IDs restart from 0 per image); cross-frame identity tracking is not implemented. ``Tracking'' behavior in annotation logic should therefore be described as heuristic and potentially incorrect if it merges identities. In addition, schema validation is structural: bounding box coordinates may be non-negative and confidence may fall in $[0,1]$, while geometric constraints (e.g., \texttt{x\_max $\le$ width}, \texttt{x\_min $<$ x\_max}) are requested by prompt but not guaranteed by validation. Finally, the interactive single-frame endpoint uses \textit{Llama-4-Scout} for qualitative inspection; it returns free-form text rather than schema-enforced JSON, and should not be described as producing production-grade structured artifacts.


\section{Background}

This project depends on one core idea: treat images like token sequences, then reuse the Transformer’s attention mechanism to integrate text instructions with visual evidence. The background below is restricted to concepts directly needed to understand the two chosen models and their system implications.

\subsection{Transformer Attention}

Transformers represent inputs as token sequences and apply self-attention to build contextual representations \cite{vaswani2017attention}. The key operation is scaled dot-product attention:

\begin{equation}
\mathrm{Attention}(Q,K,V) = \mathrm{softmax}\!\left(\frac{QK^\top}{\sqrt{d}}\right)V
\end{equation}

Here, $Q$, $K$, and $V$ are learned projections of token embeddings and $d$ is the key/query dimensionality. In multimodal settings, the ``token sequence'' may include both text tokens and visual tokens; attention then becomes the fusion mechanism that allows a generated output token to depend jointly on the prompt and the image.

\subsection{Visual Tokenization and Multimodal Fusion}

Most VLMs convert an image into a sequence of embeddings using a vision backbone, often ViT-style patch tokenization \cite{dosovitskiy2021vit}. These visual tokens are then aligned to the language model’s embedding space and inserted into the context window so the decoder can attend across a unified sequence. This framing matters for localization-like tasks: a bounding box is ultimately a function of how well spatial information is preserved in the visual tokens and how reliably the decoder can translate that spatial evidence into pixel-coordinate outputs.

\subsection{Instruction Following and Structured Generation}

Instruction-tuned models are trained to follow prompts and produce controlled outputs \cite{liu2023visual_instruction_tuning}. In engineering pipelines, a common pattern is ``structured generation'': prompts specify an explicit output contract (e.g., JSON) and downstream code validates structure \cite{openai_structured_outputs_guide}. This yields two important distinctions:
\begin{itemize}[leftmargin=*]
    \item \textbf{Structural correctness} (parsable JSON, required fields present, numeric ranges respected) can be enforced by schema validation \cite{pydantic_docs_overview}.
    \item \textbf{Semantic correctness} (boxes correspond to people; attributes reflect visual evidence) cannot be proven by schema validation and remains model-dependent.
\end{itemize}

\subsection{Mixture-of-Experts (MoE)}

MoE architectures scale capacity using conditional computation: multiple expert sub-networks exist, but only a subset are activated per token \cite{fedus2022switch_transformers}. A conceptual MoE layer can be expressed as:

\begin{equation}
y = \sum_{i \in \mathrm{TopK}} g_i(x)\, f_i(x),
\end{equation}

where $f_i$ are experts and $g_i(x)$ are routing weights produced by a gating function. MoE is relevant here because Llama-4-Scout is documented as an MoE-based model in deployment references, positioning it as an efficient choice for interactive usage at its scale \cite{nvidia_llama4_scout_nim}.

\section{Model Descriptions}

\subsection{Qwen2.5-VL-7B-Instruct}

Qwen2.5-VL-7B-Instruct is an instruction-tuned vision-language model released by the Qwen team and distributed via a public model card \cite{qwen2_5_vl_model_card}. It accepts multimodal inputs (text + image) and generates text outputs that can be natural language or structured formats when prompted appropriately. Architecturally, it follows a widely used VLM pattern:

\begin{enumerate}[leftmargin=*]
    \item \textbf{Vision encoder (image $\rightarrow$ visual tokens):} An image is converted into a sequence of embeddings, commonly using ViT-style patch representations processed with self-attention \cite{dosovitskiy2021vit}. The model card highlights support for dynamic resolution processing, which matters in practice when frames vary in resolution/aspect ratio \cite{qwen2_5_vl_model_card}.
    \item \textbf{Multimodal alignment (visual tokens $\rightarrow$ decoder token space):} Visual embeddings are mapped into a representation space that the language decoder can consume, using a projection/adapter and multimodal formatting conventions (e.g., special tokens and ordering) \cite{qwen2_5_vl_model_card}.
    \item \textbf{Transformer decoder (visual + text tokens $\rightarrow$ output tokens):} A Transformer-based decoder generates output tokens autoregressively, attending over both the instruction text and visual tokens \cite{vaswani2017attention}. This attention-based fusion is what enables grounded outputs: the model can, in principle, condition a JSON field on a specific region represented by visual tokens rather than on text priors alone.
\end{enumerate}


\needspace{7\baselineskip}
\textbf{How Qwen’s architecture maps to box outputs.} In classical object detection, bounding boxes are produced by a dedicated detection head trained directly for localization. In contrast, Qwen2.5-VL can be asked to \emph{generate} pixel-coordinate boxes as part of a structured response. This is powerful (it unifies ``detection + attribute reasoning'' under one prompt) but also fragile: the model is generating coordinates as tokens. As a result, even if the output is structurally correct JSON, boxes can be imprecise, off-by-constant, or inconsistent under occlusion/motion blur. In a defensible report, Qwen should therefore be described as a grounded generator that \emph{can be prompted} to emit boxes, not as a detector that guarantees detector-grade localization.

\textbf{Why Qwen is used for batch artifacts in the project.} The system’s batch path needs an interface contract suitable for storage, downstream aggregation, and annotation. Qwen2.5-VL is a practical fit because instruction tuning supports compliance with strict formatting constraints (field names, coordinate format, confidence fields), enabling schema-oriented prompting and downstream structural validation \cite{qwen2_5_vl_model_card,openai_structured_outputs_guide}. In the BodyLanguageDetection repository \cite{bodylanguagedetection_repo}, Qwen is prompted to detect all visible people per frame and return per-person bounding boxes and attributes (emotion by default). The architectural tradeoff is clear: one model can do perception + reasoning + structured output, but correctness must be treated probabilistically rather than guaranteed.

\subsection{Llama-4-Scout-17B-16E-Instruct}

Meta-llama/Llama-4-Scout-17B-16E-Instruct is documented in deployment references as a multimodal, instruction-tuned model and is described as incorporating Mixture-of-Experts conditional computation \cite{nvidia_llama4_scout_nim,google_vertex_llama4_scout}. Because full architectural details are not publicly disclosed, this discussion is limited to documented characteristics and standard multimodal Transformer patterns, avoiding claims that cannot be sourced.


\needspace{8\baselineskip}
\textbf{Multimodal grounding.} Like other multimodal Transformers, Scout-style deployments typically encode an image into token-level representations and place those tokens into the context window so that attention can integrate visual evidence with the instruction prompt \cite{vaswani2017attention}. This architectural principle supports interactive tasks such as ``explain what you see'' or ``describe visible cues'' because generated text can be conditioned on both the prompt and image tokens.

\textbf{MoE and interactive efficiency.} The primary architectural reason to discuss Scout in this project is MoE. Under the MoE framing, only a subset of experts are activated per token, which can improve efficiency at a given capacity scale \cite{fedus2022switch_transformers}. This makes Scout a plausible choice for interactive, developer-facing inspection where turnaround time matters.

\textbf{Role in the project: interactive qualitative inspection.} In the current implementation context \cite{bodylanguagedetection_repo}, Scout is used for a single-frame endpoint intended for prompt iteration and qualitative debugging. The endpoint returns free-form text rather than schema-enforced JSON and does not serve as the structured extraction engine. This separation of roles is a system decision grounded in model behavior: structured generation is brittle to formatting drift and validation requirements, while free-form interactive responses are useful for rapid understanding and iteration.

\needspace{110\baselineskip}

\section{Model Selection Rationale}
\FloatBarrier
\needspace{8\baselineskip}
\begingroup
\small
\renewcommand{\arraystretch}{1.05}

\captionsetup{type=table}
\captionof{table}{Comparison of selected models with respect to project requirements and architectural implications.}
\label{tab:model_comparison}

\begin{tabularx}{1.1\textwidth}{p{2.3cm} Y Y Y}
\toprule
\textbf{Requirement} &
\textbf{Qwen2.5-VL-7B-Instruct} &
\textbf{Llama-4-Scout-17B-16E-Instruct} &
\textbf{System-Level Rationale} \\
\midrule
Structured artifact generation &
Strong fit: instruction-tuned VLM that can be prompted to emit strict JSON fields for batch artifacts \cite{qwen2_5_vl_model_card,openai_structured_outputs_guide}. &
Not used as a structured artifact generator in the current interactive endpoint; returns qualitative free-form text in typical usage context. &
Batch extraction prioritizes machine-consumable artifacts; interactive inspection prioritizes human readability and iteration speed. \\
\midrule
Pixel-space bounding boxes &
Feasible via generation: can output pixel-coordinate boxes as tokens when prompted; accuracy remains model-dependent \cite{qwen2_5_vl_model_card}. &
Better suited to qualitative spatial descriptions; explicit box generation is not the focus of the interactive path. &
Bounding-box overlays require per-frame box outputs, motivating Qwen in the batch path while keeping Scout for debugging. \\
\midrule
Interactive prompt iteration &
Batch usage is higher overhead due to multi-frame processing and strict formatting constraints. &
Documented MoE framing positions it as efficient for interactive usage \cite{nvidia_llama4_scout_nim,fedus2022switch_transformers}. &
The system separates roles to avoid coupling experimentation with the structured artifact interface. \\
\midrule
Grounding mechanism &
Unified multimodal token sequence fused by attention \cite{vaswani2017attention,dosovitskiy2021vit}. &
Multimodal attention-based grounding in deployment descriptions; model-specific internals not asserted beyond documentation \cite{nvidia_llama4_scout_nim,google_vertex_llama4_scout}. &
Both models satisfy multimodal reasoning; the differentiator is output contract strictness and operational role. \\
\midrule
Failure characteristics &
Can produce valid JSON with incorrect semantics; structured formatting can drift \cite{sahoo2024hallucination_survey}. &
Free-form text avoids JSON parsing failures but is less directly consumable as an artifact. &
Structured generation improves integration when it works; interactive free-form output improves developer velocity and interpretability. \\
\bottomrule
\end{tabularx}
\endgroup

\section{Implementation Notes}

This section documents only the implementation behaviors required to keep architectural claims honest and system interfaces defensible.

\subsection{Inference Endpoints}

The batch path invokes Qwen2.5-VL through a hosted chat-completion style endpoint (e.g., via Hugging Face inference providers / OpenAI-compatible interfaces) \cite{hf_chat_completion_docs,hf_hub_inferenceclient_openai_compatible}. Frames are sampled and processed in small batches to control payload size and reduce service-limit failures.

The interactive endpoint invokes Llama-4-Scout on a single frame and returns a short free-form response. This endpoint should be described as a developer tool for qualitative inspection rather than a structured extraction component.

\subsection{Prompting Contract and Attribute Scope}

The prompt defines the output contract. In batch inference, the instruction requests:
\begin{itemize}[leftmargin=*]
    \item detection of all visible people in each frame,
    \item per-person pixel-space bounding boxes and confidence,
    \item and prompt-conditioned attributes stored in a flexible field (emotion by default).
\end{itemize}
Because attributes are prompt-conditioned, the system must not claim a fixed taxonomy unless it is explicitly enforced. For example, ``emotion'' labels may be expected in the default configuration, but additional keys in an \texttt{analysis\_result}-style dictionary are fundamentally task-dependent and determined by user.

\subsection{Schema Validation: What It Guarantees and What It Does Not}

Downstream validation checks structural properties (field presence, types, basic numeric ranges) using a schema validator \cite{pydantic_docs_overview}. This improves robustness of parsing and downstream code. However, schema validation does not prove correctness of detections:
\begin{itemize}[leftmargin=*]
    \item A JSON object can be valid while boxes do not tightly match a person.
    \item Prompt-requested geometric constraints (e.g., box inside image bounds, \texttt{x\_min < x\_max}) may not be enforced unless explicitly validated.
\end{itemize}
A defensible system therefore treats schema validation as a \emph{necessary interface layer}, not as correctness assurance.

\subsection{Identity and Temporal Consistency}

In the current prompting contract, \texttt{person\_id} values are frame-local (IDs restart per image). Cross-frame identity tracking (re-identification, temporal matching, consistent IDs across time) is not implemented and must not be implied. Any annotation logic that appears to ``track'' identities should be described as heuristic and potentially incorrect if it merges different people under the same ID.

\subsection{Failure Handling and Partial Artifacts}

Hosted inference introduces failure modes: invalid JSON, rate limits, and provider errors \cite{hf_chat_completion_docs}. A robust report should acknowledge that batches may fail and that the system may continue without producing outputs for every sampled frame depending on error handling. This is not a minor detail: it affects whether the pipeline can be described as exhaustive versus best-effort.

\section{Limitations and Risks}

This section summarizes limitations that follow from the architectural reality of generative VLMs and the system’s reliance on structured generation.

\subsection{Detection Reliability and Hallucination}

Generative models can hallucinate entities or attributes, particularly when prompts implicitly pressure the model to always answer \cite{sahoo2024hallucination_survey}. In this project context, the main failure modes are:
\begin{itemize}[leftmargin=*]
    \item \textbf{False positives/negatives:} missing a partially occluded person or inventing a person-like region.
    \item \textbf{Localization drift:} boxes are plausible but offset or inconsistently sized across frames.
    \item \textbf{Attribute overreach:} assigning an emotion label or ``cue'' not strongly supported by the pixels.
\end{itemize}
These are architecture-level limitations: a model can be excellent at producing coherent outputs while still being wrong.

\subsection{Prompt Sensitivity and Specification Drift}

Prompt wording changes model behavior \cite{salinas2024butterfly_effect_prompts}. For structured extraction, the risk is specification drift: a slight change in phrasing can change which cues are prioritized, the granularity of attributes, or whether uncertainty is expressed. Mitigation requires prompt templates, versioning, and conservative ``unknown'' policies rather than relying on ad hoc prompts.

\subsection{Privacy and Ethical Considerations}

Emotion-oriented inference from video is sensitive and can be harmful if presented as ground truth \cite{mohammad2022ethics_sheet_aer}. Risks include non-consensual analysis, re-identification concerns, and misuse of inferred attributes. When frames are transmitted to third-party inference providers, the data-handling and retention implications must be treated as first-class design constraints, not an afterthought.

\subsection{Operational Constraints}

Operational constraints include payload/token limits, rate limiting, cost variability, and external dependency risk \cite{hf_chat_completion_docs}. More frequent sampling and richer attribute dictionaries increase token usage and cost. Strict output contracts reduce downstream brittleness but can increase failure rates if the model drifts from the format. Practical deployments must therefore balance sampling rate, output verbosity, retries, and user expectations about partial failures.

\section{Conclusion}

This report adopted an architecture-led approach, using model design principles to motivate and justify system-level claims. Both selected models share a common multimodal foundation in which images are encoded as visual tokens and fused with instruction text through Transformer attention mechanisms \cite{vaswani2017attention,dosovitskiy2021vit}. Qwen2.5-VL-7B-Instruct follows the standard vision-encoder, multimodal-alignment, and Transformer-decoder paradigm and is therefore well suited for batch artifact generation, where compliance with strict JSON output formats enables reliable downstream parsing and visualization \cite{qwen2_5_vl_model_card,openai_structured_outputs_guide}. Llama-4-Scout-17B-16E-Instruct, documented as a Mixture-of-Experts multimodal model, is used to support efficient interactive analysis and prompt iteration through qualitative, free-form responses \cite{nvidia_llama4_scout_nim,fedus2022switch_transformers}.

The primary outcome of this architecture-led perspective is clarity about system guarantees. Within the implemented system context \cite{bodylanguagedetection_repo}, schema validation enforces structural correctness and improves integration reliability, but it does not ensure semantic accuracy of detections or attributes.

In summary, the system design reflects a deliberate alignment between model capabilities, interface contracts, and enforced constraints. Making these boundaries explicit provides a sound basis for technical credibility and responsible deployment.
\bibliographystyle{IEEEtran}
\bibliography{references}
\end{document}